\documentclass[7pt, conference, compsocconf]{IEEEtran}
\ifCLASSINFOpdf
   \usepackage[pdftex]{graphicx}
    \usepackage{color}
   \usepackage{pbox}
   \usepackage{multicol}
   \usepackage{multirow}
   \usepackage{pbox}
   \usepackage{hhline}
   \usepackage{setspace}

\else
\fi
\begin{document}
%
\title{CNN based Extraction of Panels/Characters from  {\em Bengali} Comic Book Page Images}
\author{\IEEEauthorblockN{Arpita Dutta, Samit Biswas}
\IEEEauthorblockA{Department of Computer Science And Technology\\
Indian Institute Of Engineering Science and Technology, Shibpur\\
Email: arpita\_dutta.rs2018@cs.iiests.ac.in, samit@cs.iiests.ac.in\\
}}



\maketitle

\begin{abstract}
Peoples nowadays prefer to use digital gadgets like cameras or mobile phones for capturing documents.  Automatic extraction of panels/characters from the images of a comic document is challenging due to the wide variety of drawing styles adopted by writers, beneficial for readers to read them on mobile devices at any time and useful for automatic digitization. Most of the methods for localization of panel/character rely on the connected component analysis or page background mask and are applicable only for a limited comic dataset.  This work proposes a panel/character localization architecture based on the features of YOLO and CNN for extraction of both panels and characters from comic book images. The method achieved remarkable results on {\em Bengali} Comic Book Image dataset (BCBId) consisting of total $4130$ images, developed by us as well as on a variety of publicly available comic datasets in other languages, i.e. eBDtheque, Manga $109$ and DCM dataset.
\end{abstract}


%
\IEEEpeerreviewmaketitle

\section{Introduction}
\begin{figure}
\begin{center}
\begin{tabular}{c}
\fbox{\includegraphics[width=3.5cm,height=4.2cm]{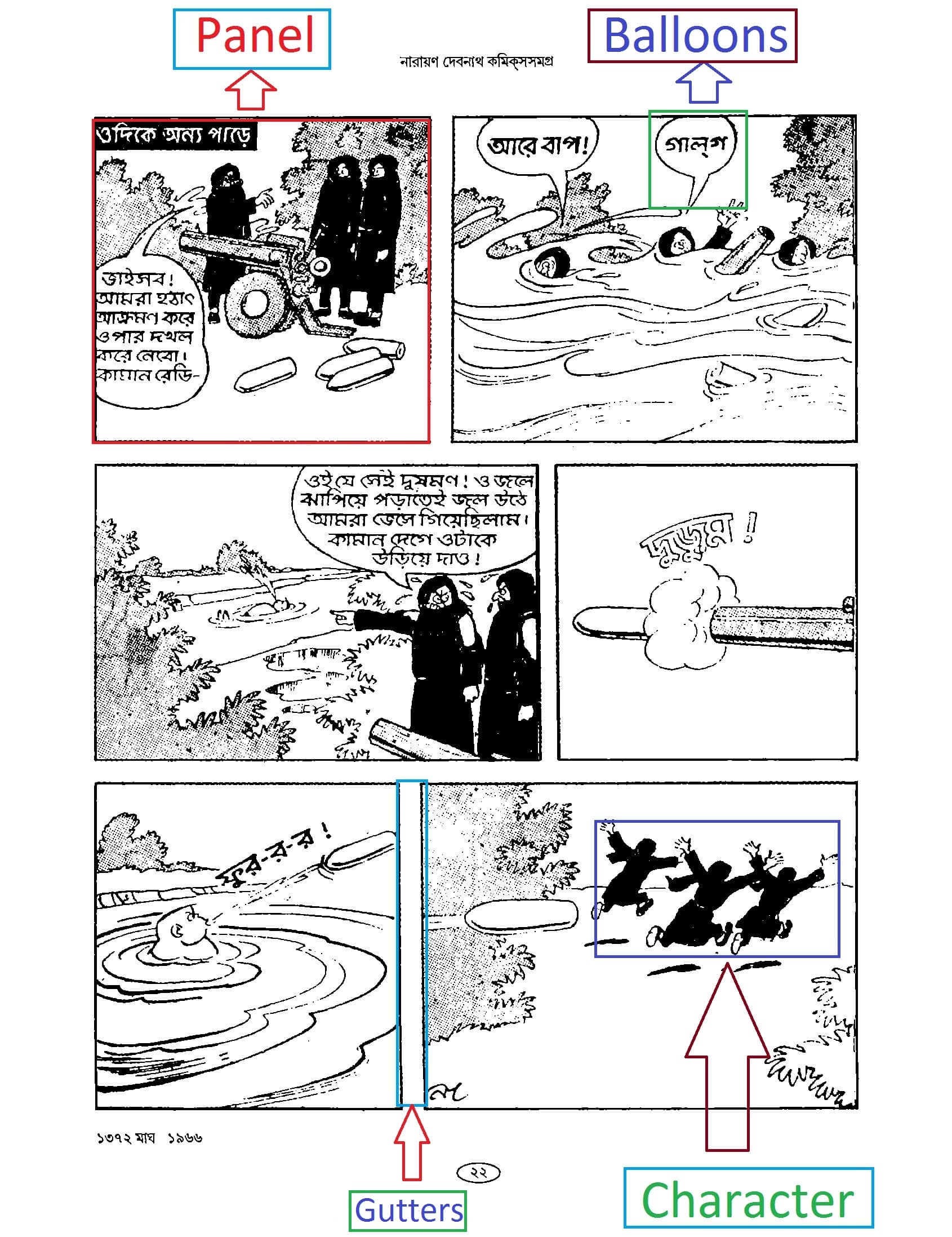}} \\
\end{tabular}
\caption{Different elements in a {\em Bengali} comic book page image. }
\label{seg}
\end{center}
\end{figure}

In our society, comics have widely used as a persuasive way to transfer information. Comics are one of the mediums for conveying ideas, expressions and concepts aesthetically; mostly preferred by children.
The Comic books are graphic novels, tell the entire story using images and texts. Nowadays, the demand for reading digital comics on mobile devices and tablets is increasing day by day with the significant improvement of technologies. There is a need to develop a well-organised system to produce digital comic from hard copies to benefit readers as most of the comic books are still in hard copies. Typically a comic book page contains several elements such as panels, texts, dialogue balloons, characters and gutters (see Figure \ref{seg}). One of the essential prerequisites for reading comics within the limited display area of mobile devices is the correct extraction of panels from the image of comic book pages. 

The comic artists incorporate a wide variety of drawing style, writing style and layout structures depending on the genre of comics.  As a result, the extraction of different components (i.e. panels, characters or balloons) for retrieval and processing of information from comic books are very arduous. Till of now, very few research works have been done related to comic book page images due to its huge variations in layout structures. Another constraint is lack of publicly available dataset; only few datasets i.e. eBDtheque\cite{guerin2013ebdtheque}, COMICS\cite{iyyer2017amazing}, Fahad $18$ dataset\cite{khan2012color} etc. are available. The structure of panels varies a lot from comics to comics. Variety of panel structures such as regular, irregular or closed panels are possible (i.e. Japanese comic Manga) \cite{pang2014robust}. {Most of the past related work do not perform well for all types of comics as those are for specific comic document images}.

{Tanaka et al.\cite{tanaka2007layout} has introduced the layout of comic pages as a sequence of panels/frames using the division lines. The algorithm determines the reading order of panels by tree traversal. The performance of the algorithm is not well in dealing with overlapping panels and other than the white background. Arai et al.\cite{arai2010automatic} located up-down and left-right overlapping panels by analyzing blobs and combining division-lines to determine frame position. Ho et al.\cite{ho2012panel}  solved the issues of overlapping frames by morphological dilation and erosion repeatedly $N$ times until it reaches the threshold value. Rigaud et al.\cite{rigaud2011robust} determined the bounding boxes of connected components and then classified them into frame, text and noise based on their features. 
Pang et al.\cite{pang2014robust} presented a method to find lines based on histogram analysis and applied a local search method to identify proper corners of overlapping panels. Wang et al.\cite{wang2015comic} identified frame polygons based on labelling of connected components and a combination of line segments. The proposed algorithm analyzes lines of each component and then tries to optimize a cost function by several rules to locate panels. Most of the existing algorithms for panel detection rely on the connected component analysis or division-line for panel detection.  Therefore, those methods do not perform well in case of complex panel layout or the absence of white margin. Iyyer et al.\cite{iyyer2017amazing} proposed a deep learning-based method for extraction of panels from comic books.

{On the other hand, comic character detection is another essential step in different applications of comics. However, it is another challenging task to deal with due to its several constraints in comparison with faces of real images. The constraints are: 1) There exists a significant difference among shapes, sizes, positions of organ and expressions of comic characters. 2) Generally, comic characters are drawn using straight lines or curved lines and have very little colour information. Moreover, there exists a considerable variety among the number of characters across comic images. As a result, the existing algorithms for human face detection in real images based on classical approaches cannot efficiently tackle comic character detection challenges. Rigaud et al.\cite{rigaud2014color} proposed a method to retrieve a comic character from colour comic books based on the colour descriptor. In another work, Rigaud et al.\cite{rigaud2015knowledge} introduced a method to localize comic characters based on analysis of speech balloon components in an unsupervised way. Sun et al.\cite{sun2013specific} proposed a method based on the similarity of different features, i.e. poses, expression etc. for comic character identification.}

{To our knowledge, recently, Qin et al.\cite{qin2017faster} and Nguyen et al.\cite{nguyen2017comic} have used directly Faster R-CNN\cite{ren2015faster} and YOLOv2 model \cite{redmon2017yolo9000} respectively for comic character identification. However, in this work, instead of directly using any particular method, our proposed method has amalgamated different features of YOLO algorithm \cite{redmon2016you,redmon2017yolo9000,redmon2018yolov3} based on our research problems. This work proposes a CNN based method for extraction of both panels and characters simultaneously from comic book images. The main contribution of this work is as follows: Firstly, we introduce a new dataset on Bengali Comic book pages which includes various kinds of structural layouts and huge variations of comic characters. We annotate the entire character because only face regions of comic character are not distinguishable enough to identify characters further \cite{sun2010similar}}. {Secondly, proposing a panel/character localization architecture based on the features of YOLO and CNN. It directly predicts the location of panels and characters from the input comic book page image.} 

The rest of the paper is organised as follows:  the proposed method in brief is described in Section-\ref{PropMethod}, Section-\ref{expResult} shows experimental results and finally, we conclude in Section-\ref{ConRemark}.

\section{Proposed Method} \label{PropMethod}
The use of Convolutional Neural Network in computer vision tasks has increased from the last few years \cite{krizhevsky2012imagenet}. To our knowledge, very few works have used deep learning methods for comic document image analysis \cite{qin2017faster,nguyen2017comic,iyyer2017amazing}. Here, the objective is to segment panels/characters from comic book images. Therefore, the task is to detect bounding box across the objects (panels/characters) and assign it to its appropriate label (panel/ character). As the problem is related to design a predictive model (both object localization and classification), this work uses Convolutional Neural Networks (CNN). Here, the CNN estimates the coordinates of bounding box across objects by minimizing the error between the detected bounding box and the ground truth bounding box and simultaneously calculates class probabilities of those detected bounding boxes directly from image pixels.

\subsection{Deep learning based object detection}
The aim of most of the research methods based on region proposal \cite{girshick2014rich,he2015spatial,girshick2015fast} is to reduce the number of regions for classification and to improve the accuracy of localization.  In \cite{girshick2015fast}, the author combines classification and localization network to reduce the training time and to increase accuracy. The Faster R-CNN \cite{ren2015faster} for object detection determines region proposals using a deep neural network and utilizes it instead of handcrafted features to obtain high accuracy.

Redmon et al.\cite{redmon2016you} proposed a method named YOLO where bounding box coordinates are directly predicted from input images and objects are assigned to its proper class. This algorithm can detect objects with high speed in a real-time system without sacrificing too much accuracy. Later, the same author proposed another two modified approaches \cite{redmon2017yolo9000, redmon2018yolov3} of YOLO to tackle complex object detection problem. Next subsection shows an overview of YOLO algorithm.
\subsection{YOLO Algorithm}
The idea behind YOLO algorithm is to split the entire image first into $S \times S$ grid cells. Then, the algorithm will estimate B number of bounding boxes for each grid cell. Now, each bounding box, say $B_{j}$ has following predictions: X, Y, W, H, $P_{obj}$ and $P_{c}$. Here, (X,Y) denotes the centre of bounding box, W and H signify width and height respectively, $P_{obj}$ denotes the class probability such that an object is located at the centre inside the cell and lastly, $P_{c}$ denotes the conditional probability of class c for the bounding box $B_{j}$, where $c \in L$, L is the set of all target classes to be detected. In YOLO \cite{redmon2016you}, the locations of bounding boxes are directly predicted using fully convolutional neural network where as in YOLO9000 \cite{redmon2017yolo9000} hand-crafted priors are used to predict the coordinates of bounding boxes. 

\begin{figure*}
	\begin{center}
		\begin{tabular}{c}			
			\fbox{\includegraphics[scale=0.34]{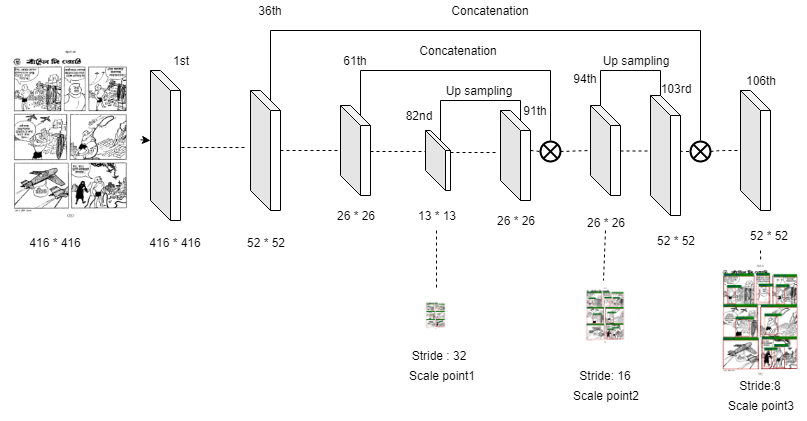}}\\			
		\end{tabular}		
		\caption{CNN architecture for panel and character localization}
		\label{Archi}
	\end{center}
\end{figure*}

\subsection{ Panel/Character Localization Architecture}
 YOLO algorithm \cite{redmon2017yolo9000} is used to tackle real time object detection problem; it uses softmax classifier for detection of several kinds of objects. The probability of target class estimated by softmax classifier can be given by:

 \begin{equation} 
		\sigma _{j} (y) = {\frac{\exp (y_{j})}{\sum_{i=1}^{c} \exp (y_{i})}},   j={1,2,.....,c}
		\label{eq128} 		
\end{equation}
Where $\sigma _{j} (y)$ denotes probability estimates, $y$ denotes the output taken from last fully connected layer, $c$ is the total number of target class, and $y_{j}$ corresponds to the $j^{th}$ class. However, softmax classifier is a generic classifier for multilabel classification problem. That is why, we need to deal with huge number of parameters during implementation of softmax classifier in deep learning framework. However, sigmoid classifier performs well in case of binary classification problem. Softmax classifier is actually generalised version of sigmoid classifier (in later, the number of target class is two). The probability estimations calculated by sigmoid function is defined as follows: $\sigma (y) = {\frac{1}{1 + e^{-y}}} $

This work uses sigmoid classifier instead of softmax classifier as there are two target classes (Panel and Character). It deals with very fewer parameters and it smooths the optimization process. To demonstrate the superiority of sigmoid classifier over softmax classifier, we have trained our method on the proposed dataset using both softmax and sigmoid classifier. Moreover, we have also trained YOLO\cite{redmon2016you} and YOLO9000 \cite{redmon2017yolo9000} on our proposed dataset. Table- \ref{tab:xxxloss} shows the corresponding training and cross validation loss using original YOLO\cite{redmon2016you}, YOLO9000 \cite{redmon2016you}, the proposed method using softmax and sigmoid classifier. It is clearly visible that the proposed method with sigmoid classifier has achieved the minimized loss compared to all other method. It also helps to train the network for a limited dataset. But, the proposed method is not limited to two class classification problem only. If we replace sigmoid classifier with softmax classifier, then the proposed architecture can be easily used for multiclass classification problem. In our architecture, we have made detection at three different scale in the network. As a result, it can efficiently detect small objects by concatenating features from previous layers.

\begin{table}[h]

\caption{Cross Validation Loss during training }
\centering
        \tiny
        
        {\begin{tabular}{|c|c|c|c|}
        \hline
        \pbox{2.2cm}{\bf Method} & {\bf Dataset} & \pbox{2.2cm}{\bf Traning} & \pbox{3.5cm}{\bf Cross-validation}  \\
                     &              &\pbox{2.2cm}{\bf Loss}                         & \pbox{3.6cm}{\bf Loss}  \\
            \hline
            
          YOLO \cite{redmon2016you} & {\em BCBId} & {2.4700000}& {3.1075} \\
                                    \hline
 YOLO9000 \cite{redmon2017yolo9000} & { \em BCBId} & 1.8056001 & 2.4205 \\
 \hline

\ Our method (Softmax)  & { \em BCBId} & 0.7200001 & 1.2005 \\
\hline

\ Our method (Sigmoid)  & { \em BCBId} & 0.0000001 & 0.0007 \\
\hline

        \end{tabular}}{}
        
\label{tab:xxxloss}

\end{table}

In YOLO9000 \cite{redmon2017yolo9000}, Darknet-$19$ model, which consists of $19$ convolutional layers along with $5$ max-pooling layers is used for object detection. But it fails to detect small objects due to generation of loss for downsampling. To overcome this problem, our network takes output at three different scale and incorporates the idea of residual skip connections and upsampling while designing network. In this work, we have stacked two Darknet-53 \cite{redmon2018yolov3}, resulting total $106$ layers. Though, the number of layers have been increased, but it does not increase the number of parameters as it uses sigmoid classifier. We have applied $1 \times 1$ kernel on feature maps of various sizes at three different positions in the network.

The output shape of detection kernel can be defined as {$1 \times 1 \times (B \times (5 + L))$}, where $B$ denotes the number of bounding box predicted by each grid cell on the feature map, $L$ is the total number of classes and $5$ signifies the predictions (X, Y, W, H and $P_{obj}$ ) by each bounding boxes as mentioned earlier. In our work, as there are two target labels (Panel and Character), here the value of $L$ will be $2$. We have taken $3$ bounding box for each grid cell, so the value of $B$ will be $3$. Therefore, the size of kernel is $1 \times 1 \times 21 $.




In our proposed dataset, the resolution of input image is high (see Section \ref{expResult}). Therefore, in our method, we scale down the input image size, feed it to the network to predict co-ordinates of bounding boxes and then again scale up to its original size. The size of input image fed to the network is $416 \times 416$. As, the prediction is done at three different scale, so the resolutions of input images are downsampled by a factor of $8$, $16$ and $32$ respectively (see figure \ref{Archi}). The output of first detection is taken from $82^{nd}$ layer. The dimension of the input image is down sampled by the network in the previous $81$ layers, such that the value of stride at $81^{st}$ layer is $32$. As, the input image size is $416 \times 416$, then the size of feature map at $82^{nd}$ layer will be $13 \times 13$. Again, as the detection is done using $1 \times 1$ detection kernel, therefore, the final shape of detection feature map obtained from $82^{nd}$ layer is $13 \times 13 \times 21$. {The feature map size at $61^{st}$ layer is $26 \times 26$, it has stride of size $16$. The output obtained from $82^{nd}$ layer is  then upsampled such that at $91^{th}$ layer, the size of feature map will be $26 \times 26$. After that, the feature map obtained from $61^{th}$ layer is concatenated with the feature map obtained from $91^{th}$ layer. Then again few $1 \times 1$ convolution layers are added on this combined feature maps to amalgamate the features from previous layer i.e $61^{st}$ layer. Then finally, the second detection is taken from $94^{th}$ layer, the final output shape of detection feature map is $26 \times 26 \times 21$. Using the similar approach, the feature map obtained from $94^{th}$ layer is upsampled again such that at layer $103^{rd}$, the size of the feature map will be $52 \times 52$. On the other hand, the size of feature map obtained from $36^{th}$ layer is $52 \times 52$ with stride of size $8$. In the next step, the feature map obtained from $36^{th}$ layer is  concatenated with the feature map from layer $103^{rd}$. Like in previous case, here also few $1 \times 1$ convolution layers are added on this concatenated feature map to merge the information from earlier layer i.e $36^{th}$ layer.} Then, we finally predict the third detection at $106^{th}$ layer with a detection feature map of size $52 \times 52 \times 21$. The detection made at scale $point_1$ i.e $13 \times 13$ layer can successfully identify large objects. On the other hand, the scale $point_3$ i.e $52 \times 52$ can efficiently detect small objects. However, $26 \times 26$ layer i.e scale $point_2$ is responsible for detecting medium objects. Finally, we scale up the output obtained from $106th$ layer (size $52 \times 52$) to the original image size.
However, to signify the effectiveness of amalgamating features of different scale proposed by our method over YOLO\cite{redmon2016you} and YOLO9000 \cite{redmon2017yolo9000}, we have presented the experimental result conducted on YOLO\cite{redmon2016you} and YOLO9000 \cite{redmon2017yolo9000}. The proposed method has achieved the best performance and the improvements of all precision, recall, F-measure and IoU are shown in  Table \ref{tab:xxxpanel} and Table \ref{tab:xxxcharacter}.


Like YOLO9000 \cite{redmon2017yolo9000} algorithm, here we also need prior anchors to train the localization network. As, here the number of bounding box per grid is $3$ and the output is taken from $3$ different scale, so the number of total anchors are ($3 \times 3 = 9$). We have generated those anchors using K-Means clustering algorithm and then arranged them in descending order. Then we have assigned the first three anchors for scale $point_1$, next three for scale $point_2$ and last three for scale $point_3$. However, the total number of bounding boxes are $((52 \times 52 \times 3) + (26 \times 26 \times 3) + (13 \times 13 \times 3) = 10647)$. Among those boxes, our algorithm finally chooses those bounding boxes which have maximum overlap with its corresponding ground truth boxes. To predict more accurate bounding boxes and clean up irrelevant predictions, we have used non-maximal suppression as a post processing step like the original YOLO\cite{redmon2016you}. For non-maximal suppression, we have measured objectness score and confidence score of each predicted bounding boxes. Objectness score signifies that there is an object within the bounding box. Now, if the objectness score $P_{obj}$ is greater than $70\%$, then those bounding boxes will be considered as finally predicted bounding boxes and rest will be discarded. But, the predicted bounding box may predict classes that are not in target class. Confidence score tackles this problem easily because it actually signifies how accurately the method can predict label of bounding boxes (confidence score = $P_{obj}$ $\times$ IoU of bounding boxes with ground truth boxes). we have checked the IoU of bounding boxes with ground truth boxes and which have more than $80\% $ IoU , are considered as finally detected boxes (details of IoU in experimental result section). Therefore, the finally detected boxes have at least more than 80\% IoU overlap with its corresponding ground truth boxes. The bounding boxes, which have less than 80\% IoU overlap with its corresponding ground truth box, are rejected. Therefore, we always find at most one match corresponding to each ground truth boxes. 

\subsection{Implementation Details}
Due to lack of available dataset of comic book pages with annotation, here, we have used the idea of transfer learning \cite{NIPS2014_5347}. We have used the weights pretrained on ImageNet dataset \cite{Russakovsky2015} with VGG16 \cite{DBLP:journals/corr/SimonyanZ14a} network during training of our proposed model. Then, we have done fine tuning our model on proposed dataset (mentioned in Section \ref {expResult}). We have used Adam optimizer for optimization of loss function. We have trained our network with $70,000$ iterations, and set learning rate as $0.001$ for first $42,000$ iterations and then again updated the learning rate as $0.0001$ for rest $28,000$ iterations.  

\section{Experimental Results} \label{expResult}
Since there is no standard benchmark Bengali Comic Image Database for comic document image analysis, we have created our own {\em Bengali} Comic Dataset. {Performance of this work is measured using our developed Bengali Comic Dataset along with publicly available  dataset such as eBDtheque \cite{guerin2013ebdtheque}, Manga $109$ \cite{matsui2017sketch} and DCM dataset \cite{Christop24:online}.}

\paragraph{Bengali Comic Book Image Dataset (BCBId)}\label{bangla}
We build a new dataset on {\em Bengali} comic book pages which captures huge variations in drawing styles incorporated by comic artists.
To our knowledge, there are no such existing dataset to pursue research on {\em Bengali} comic book page images. The entire dataset is categorized based on the drawing style of comic writers. Till of now, we have constructed the dataset depending on the comic books written by three writers namely Narayan Debnath, Shivram Chakrabarty and Premendra Mitra. Our dataset is made of $980$ images from comic books written by Narayan Debnath, named the {\em ND980} dataset. The dimension of images in ND980 dataset is $1690 \times 2195$. The second part of our dataset consists of $1400$ images of comic book pages written by Shivram Chakrabarty, named the SC1400 dataset. In SC1400 dataset, the image resolution is $2232 \times 3072$. Another part contains $1750$ images written by Premendra Mitra, named the PM1750 dataset. The dimension of images in this dataset is very high; it is $6250 \times 8750$. We annotate different components (balloon, panel and characters) using \textbf{VGG Annotation Tool}\cite{VGGImage12:online} so that researchers can utilize them according to their specific task.

In this work, we have taken $60\%$ images from every three dataset for traning, $20\%$ images for validation and rest $20\%$ images for testing. More specifically, we have used $588$, $196$, and $196$ images from ND980 dataset for training, validation and testing respectively. On the other hand, $840$, $280$ and $280$ images from SC1400 dataset and $1050$, $350$ and $350$ images from PM1750 dataset  are used for training, validation and testing respectively. 
\begin{figure}[h]
\begin{center}
\begin{tabular}{ll}

a) \fbox{\includegraphics[width=3.5cm,height=4.2cm]{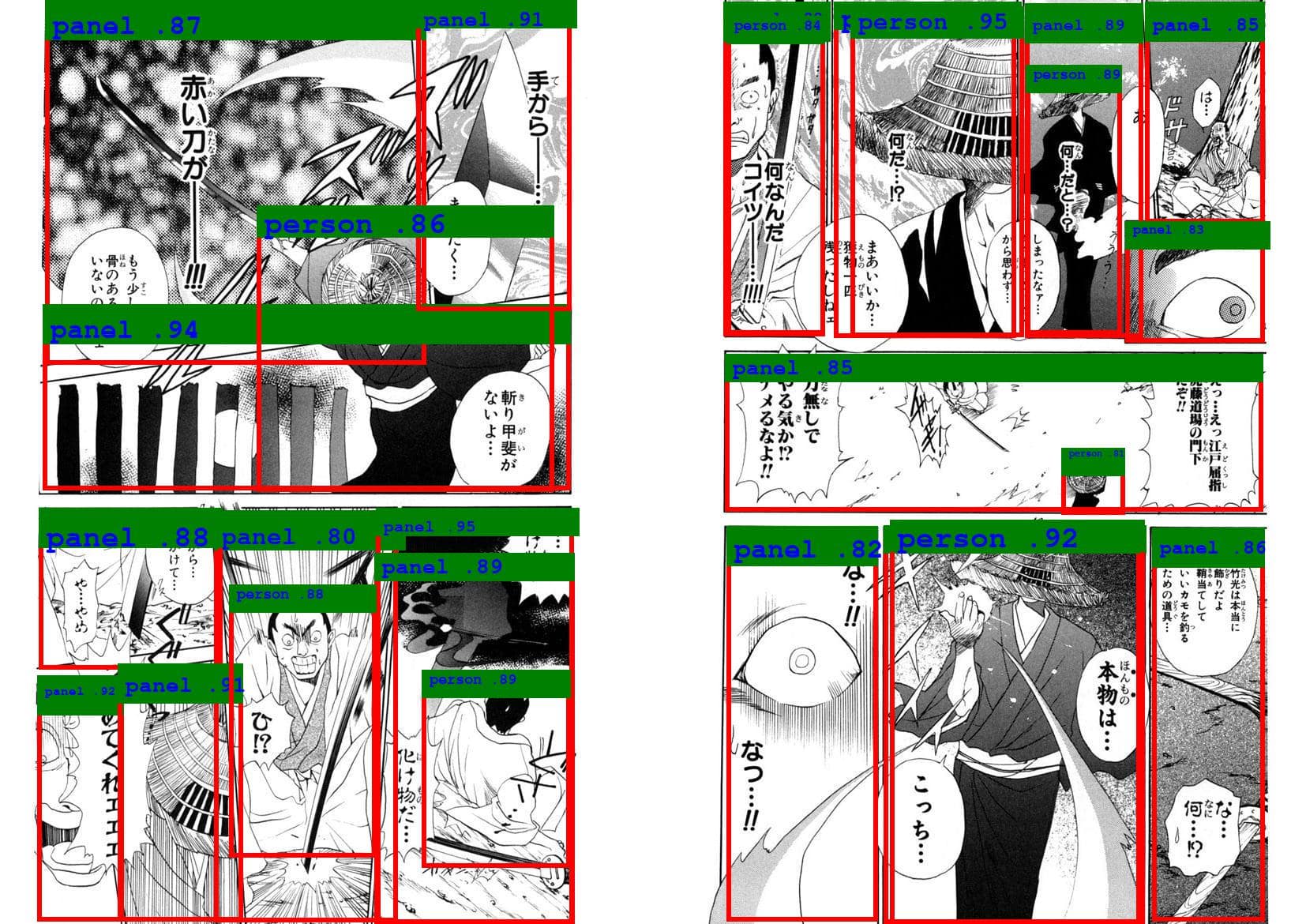}} & 
b)\fbox{\includegraphics[width=3.5cm,height=4.2cm]{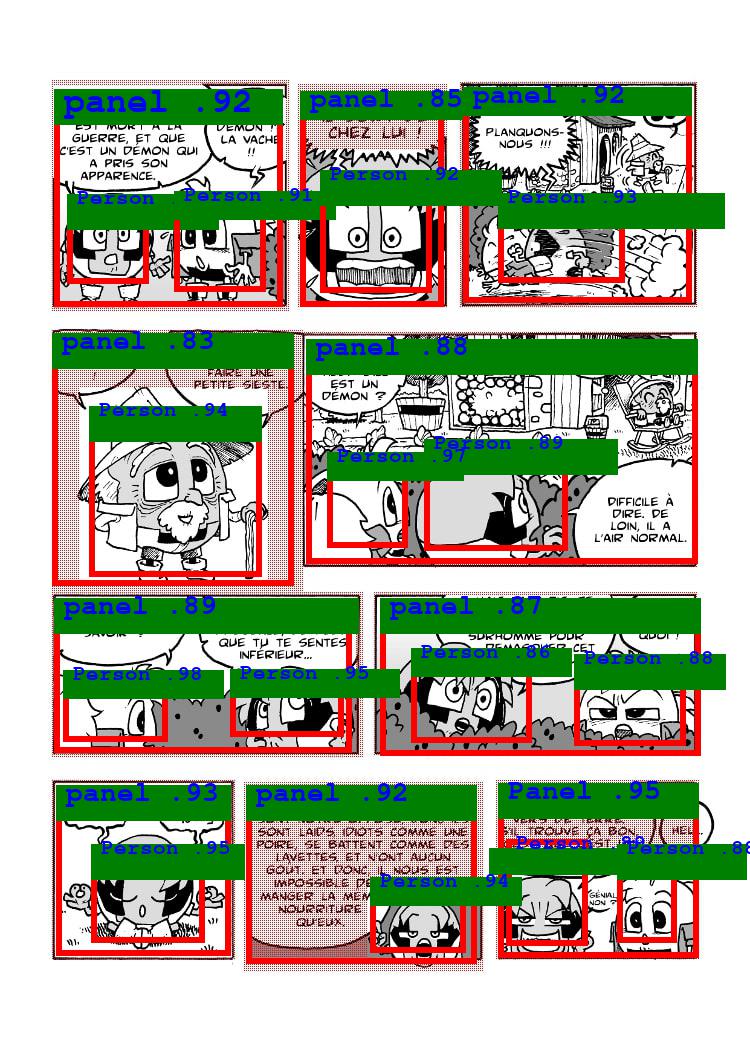}}\\

\end{tabular}
\end{center}
\caption{Result using proposed method: a) Manga $109$ dataset; b) \em {eBDtheque} dataset;}
	\label{resultmanga}
\end{figure}

\begin{table}[h]
 \caption{Evaluation of Panel detection on eBDtheque dataset \cite{guerin2013ebdtheque}, Manga $109$\cite {matsui2017sketch}, DCM \cite{Christop24:online} and {\em BCBId}.}
\centering
	\tiny

        {\begin{tabular}{|c|c|c|c|c|c|}
        \hline
        \pbox{1cm}{\bf Method} & {\bf Dataset} & \pbox{1.5cm}{\bf Preci-} & \pbox{1.5cm}{\bf Re-} & \pbox{1.5cm}{\bf F-mea-} & \pbox{2.8cm}{\bf IoU} \\
                        & \pbox{1.5cm}{\bf used}       & \pbox{1.5cm}{\bf sion}  & \pbox{1.5cm}{call} & \pbox{1.5cm}{\bf sure}  & \% \\
                        &              & \%  & \% & \%  &  \\
          \hline
          
          \multirow{2}{1cm}{Rigaud et al.\cite{rigaud2011robust}} & \multicolumn{1}{c|}{eBDtheque \cite{guerin2013ebdtheque}} & \pbox{1.5cm}{$63$} &\pbox{1.5cm}{$69$} & \pbox{1.5cm}{$66$} & \pbox{1.5cm}{63}  \\\hhline{~-----}
          
          &
          
           \multicolumn{1}{c|}{Manga $109$ \cite{matsui2017sketch}} & \pbox{1.5cm}{$70.10$} &\pbox{1.5cm}{$68.20$} & \pbox{1.5cm}{$69.13$} & \pbox{1.5cm}{70}  \\\hhline{~-----}
          
          &
          
          \multicolumn{1}{c|}{DCM \cite{Christop24:online}} & \pbox{1.5cm}{$59.70$} &\pbox{1.5cm}{$64.71$} & \pbox{1.5cm}{$62.11$} & \pbox{1.5cm}{$60$}  \\\hhline{~-----}

                                   & \multicolumn{1}{c|}{{\em BCBId}} & \pbox{1.5cm}{$56$} & \pbox{1.5cm}{$52$} & \pbox{1.5cm}{$53.92$} & \pbox{1.5cm}{$50$} \\
                    
                                   \hline

          \multirow{2}{1cm}{Wang et al.\cite{wang2015comic}} & \multicolumn{1}{c|}{eBDtheque \cite{guerin2013ebdtheque}} & \pbox{1.5cm}{$84$} &\pbox{1.5cm}{$70$} & \pbox{1.5cm}{$76$} & \pbox{1.5cm}{$70$}  \\\hhline{~-----} &
          
          \multicolumn{1}{c|}{Manga $109$ \cite{matsui2017sketch}} & \pbox{1.5cm}{$82$} &\pbox{1.5cm}{$80.51$} & \pbox{1.5cm}{$81.24$} & \pbox{1.5cm}{$81$}  \\\hhline{~-----} &
          
          \multicolumn{1}{c|}{DCM \cite{Christop24:online}} & \pbox{1.5cm}{$77.22$} &\pbox{1.5cm}{$79.14$} & \pbox{1.5cm}{$78.13$} & \pbox{1.5cm}{$77$}  \\\hhline{~-----}
          
                                   & \multicolumn{1}{c|}{{\em BCBId}} & \pbox{1.5cm}{$77$} & \pbox{1.5cm}{$63$} & \pbox{1.5cm}{$67.63$} & \pbox{1.5cm}{$65$} \\
                                   
                                   \hline
                                   
          \multirow{2}{1cm}{Pang et al.\cite{pang2014robust}} & \multicolumn{1}{c|}{eBDtheque \cite{guerin2013ebdtheque}} & \pbox{1.5cm}{$74$} &\pbox{1.5cm}{$73$} & \pbox{1.5cm}{$73.49$} & \pbox{1.5cm}{$74$}  \\\hhline{~-----} &
          \multicolumn{1}{c|}{Manga $109$ \cite{matsui2017sketch}} & \pbox{1.5cm}{$90.14$} &\pbox{1.5cm}{$92.56$} & \pbox{1.5cm}{$91.35$} & \pbox{1.5cm}{$92$}  \\\hhline{~-----} &
          
           \multicolumn{1}{c|}{DCM \cite{Christop24:online}} & \pbox{1.5cm}{$73$} &\pbox{1.5cm}{$75.28$} & \pbox{1.5cm}{$74.08$} & \pbox{1.5cm}{$75$}  \\\hhline{~-----}
                                   & \multicolumn{1}{c|}{{\em BCBId}} & \pbox{1.5cm}{$70$} & \pbox{1.5cm}{$69$} & \pbox{1.5cm}{$64.61$} & \pbox{1.5cm}{$63$} \\

                                   \hline

\multirow{2}{1cm}{Yolo. \cite{redmon2016you}} & \multicolumn{1}{c|}{eBDtheque \cite{guerin2013ebdtheque}} & \pbox{1.5cm}{$78.28$} &\pbox{1.5cm}{$76.35$} & \pbox{1.5cm}{$77.28$} & \pbox{1.5cm}{$76$}  \\\hhline{~-----} &
          \multicolumn{1}{c|}{Manga $109$ \cite{matsui2017sketch}} & \pbox{1.5cm}{$85.67$} &\pbox{1.5cm}{$83.25$} & \pbox{1.5cm}{$84.38$} & \pbox{1.5cm}{$85$}  \\\hhline{~-----} &
          
           \multicolumn{1}{c|}{DCM \cite{Christop24:online}} & \pbox{1.5cm}{$80.98$} &\pbox{1.5cm}{$73.82$} & \pbox{1.5cm}{$77.14$} & \pbox{1.5cm}{$79$}  \\\hhline{~-----}
                                   & \multicolumn{1}{c|}{{\em BCBId}} & \pbox{1.5cm}{$72.34$} & \pbox{1.5cm}{$75.97$} & \pbox{1.5cm}{$74.05$} & \pbox{1.5cm}{$75$} \\

                                   \hline

\multirow{2}{1cm}{YOLO9000. \cite{redmon2017yolo9000}} & \multicolumn{1}{c|}{eBDtheque \cite{guerin2013ebdtheque}} & \pbox{1.5cm}{$85.37$} &\pbox{1.5cm}{$84.76$} & \pbox{1.5cm}{$85$} & \pbox{1.5cm}{$86$}  \\\hhline{~-----} &
          \multicolumn{1}{c|}{Manga $109$ \cite{matsui2017sketch}} & \pbox{1.5cm}{$90.26$} &\pbox{1.5cm}{$87.83$} & \pbox{1.5cm}{$89$} & \pbox{1.5cm}{$88.58$}  \\\hhline{~-----} &
          
           \multicolumn{1}{c|}{DCM \cite{Christop24:online}} & \pbox{1.5cm}{$84.35$} &\pbox{1.5cm}{$82.17$} & \pbox{1.5cm}{$83.12$} & \pbox{1.5cm}{$84$}  \\\hhline{~-----}
                                   & \multicolumn{1}{c|}{{\em BCBId}} & \pbox{1.5cm}{$79.62$} & \pbox{1.5cm}{$78.95$} & \pbox{1.5cm}{$79.26$} & \pbox{1.5cm}{$80$} \\

                                   \hline

         \multirow{2}{1.5cm}{Our method \\(Using \\Softmax)} & \multicolumn{1}{c|}{eBDtheque \cite{guerin2013ebdtheque}} & \pbox{1.5cm}{$92.74$} &\pbox{1.5cm}{$93.28$} & \pbox{1.5cm}{$93$} & \pbox{1.5cm}{$92.56$}  \\\hhline{~-----}
         &
         \multicolumn{1}{c|}{Manga $109$ \cite{matsui2017sketch}} & \pbox{1.5cm}{$90.86$} &\pbox{1.5cm}{$91.14$} & \pbox{1.5cm}{$91$} & \pbox{1.5cm}{$90$}  \\\hhline{~-----}
         
         &
         
         \multicolumn{1}{c|}{DCM \cite{Christop24:online}} & \pbox{1.5cm}{$93.52$} &\pbox{1.5cm}{$94.37$} & \pbox{1.5cm}{$93.85$} & \pbox{1.5cm}{$94$}  \\\hhline{~-----}
                                   & \multicolumn{1}{c|}{BCBId} & \pbox{1.5cm}{$92.91$} & \pbox{1.5cm}{$94.10$} & \pbox{1.5cm}{$93.42$} & \pbox{1.5cm}{$92$} \\

                                   \hline

         \multirow{2}{1.5cm}{Our method \\(Using \\ Sigmoid)} & \multicolumn{1}{c|}{eBDtheque \cite{guerin2013ebdtheque}} & \pbox{1.5cm}{$97$} &\pbox{1.5cm}{$98$} & \pbox{1.5cm}{$97.49$} & \pbox{1.5cm}{$98$}  \\\hhline{~-----} &
         \multicolumn{1}{c|}{Manga $109$ \cite{matsui2017sketch}} & \pbox{1.5cm}{$98.76$} &\pbox{1.5cm}{$97.25$} & \pbox{1.5cm}{$97.92$} & \pbox{1.5cm}{$98$}  \\\hhline{~-----} &
         
         \multicolumn{1}{c|}{DCM \cite{Christop24:online}} & \pbox{1.5cm}{$98.28$} &\pbox{1.5cm}{$97.55$} & \pbox{1.5cm}{$97.89$} & \pbox{1.5cm}{$97$}  \\\hhline{~-----} 
         
                                   & \multicolumn{1}{c|}{{\em BCBId}} & \pbox{1.5cm}{$98.55$} & \pbox{1.5cm}{$98$} & \pbox{1.5cm}{$98.24$} & \pbox{1.5cm}{$98.56$} \\

                                   \hline

         \end{tabular}}{}
\label{tab:xxxpanel}

\end{table}

\begin{table}[h]

 \caption{Evaluation of Character detection on eBDtheque dataset\cite{guerin2013ebdtheque}, Manga $109$\cite {matsui2017sketch}, DCM \cite{Christop24:online} and {\em BCBId}.}
\centering
		\tiny
        {\begin{tabular}{|c|c|c|c|c|c|}
        \hline
        \pbox{1cm}{\bf Method} & {\bf Dataset} & \pbox{1.5cm}{\bf Preci-} & \pbox{1.5cm}{\bf Re-} & \pbox{1.5cm}{\bf F-mea-} & \pbox{2.8cm}{\bf IoU} \\
                        & \pbox{1.5cm}{\bf used}       & \pbox{1.5cm}{\bf sion}  & \pbox{1.5cm}{call} & \pbox{1.5cm}{\bf sure}  & \% \\
                        &              & \%  & \% & \%  &  \\
          \hline
          
        \multirow{2}{1.5cm}{Rigaud et al.\cite{rigaud2015knowledge}} & \multicolumn{1}{c|}{eBDtheque \cite{guerin2013ebdtheque}} & \pbox{1.5cm}{$21.57$} &\pbox{1.5cm}{$40.52$} & \pbox{1.5cm}{$28.16$} & \pbox{1.5cm}{25}  \\\hhline{~-----}
        &
        \multicolumn{1}{c|}{Manga $109$ \cite{matsui2017sketch}} & \pbox{1.5cm}{$19.14$} &\pbox{1.5cm}{$23.20$} & \pbox{1.5cm}{$21$} & \pbox{1.5cm}{$22$}  \\\hhline{~-----}
        &
        \multicolumn{1}{c|}{DCM \cite{Christop24:online}} & \pbox{1.5cm}{$28.32$} &\pbox{1.5cm}{$25.65$} & \pbox{1.5cm}{$26.80$} & \pbox{1.5cm}{$26$}  \\\hhline{~-----}
        
                                   & \multicolumn{1}{c|}{BCBId} & \pbox{1.5cm}{$17$} & \pbox{1.5cm}{$35$} & \pbox{1.5cm}{$22.84$} & \pbox{1.5cm}{$21$} \\
                    
                                   \hline
                                   
    \multirow{2}{1.5cm}{Sun et al.\cite{sun2013specific}} & \multicolumn{1}{c|}{eBDtheque \cite{guerin2013ebdtheque}} & \pbox{1.5cm}{$79.43$} &\pbox{1.5cm}{$35.48$} & \pbox{1.5cm}{$49.05$} & \pbox{1.5cm}{$50$}  \\\hhline{~-----}
    &
    \multicolumn{1}{c|}{Manga $109$ \cite{matsui2017sketch}} & \pbox{1.5cm}{$65.20$} &\pbox{1.5cm}{$70.70$} & \pbox{1.5cm}{$67.85$} & \pbox{1.5cm}{$65$}  \\\hhline{~-----}
    &
    \multicolumn{1}{c|}{DCM \cite{Christop24:online}} & \pbox{1.5cm}{$71.22$} &\pbox{1.5cm}{$63$} & \pbox{1.5cm}{$66.82$} & \pbox{1.5cm}{$67$}  \\\hhline{~-----}
                                   & \multicolumn{1}{c|}{BCBId} & \pbox{1.5cm}{$62$} & \pbox{1.5cm}{$56$} & \pbox{1.5cm}{$58.84$} & \pbox{1.5cm}{$56$} \\
                    
                                   \hline

    \multirow{2}{1.5cm}{Qin et al.\\(softmax)\cite{qin2017faster}}& \multicolumn{1}{c|}{eBDtheque \cite{guerin2013ebdtheque}} & \pbox{1.5cm}{$70.92$} &\pbox{1.5cm}{$48.41$} & \pbox{1.5cm}{$57.50$} & \pbox{1.5cm}{$56$}  \\\hhline{~-----}
    &
    \multicolumn{1}{c|}{Manga $109$ \cite{matsui2017sketch}} & \pbox{1.5cm}{$72.71$} &\pbox{1.5cm}{$65.23$} & \pbox{1.5cm}{$68.70$} & \pbox{1.5cm}{$69$}  \\\hhline{~-----}
    &
    \multicolumn{1}{c|}{DCM \cite{Christop24:online}} & \pbox{1.5cm}{$69.14$} &\pbox{1.5cm}{$72.32$} & \pbox{1.5cm}{$70.65$} & \pbox{1.5cm}{$69$}  \\\hhline{~-----}
    
                                   & \multicolumn{1}{c|}{BCBId} & \pbox{1.5cm}{$67$} & \pbox{1.5cm}{$69.26$} & \pbox{1.5cm}{$68.08$} & \pbox{1.5cm}{$65$} \\
                                   
                                   \hline
                                   
      \multirow{2}{1.5cm}{Qin et al.\\(sigmoid)\cite{qin2017faster}}& \multicolumn{1}{c|}{eBDtheque \cite{guerin2013ebdtheque}} & \pbox{1.5cm}{$75.25$} &\pbox{1.5cm}{$49.85$} & \pbox{1.5cm}{$60.10$} & \pbox{1.5cm}{$58$}  \\\hhline{~-----}
      &
      \multicolumn{1}{c|}{Manga $109$ \cite{matsui2017sketch}} & \pbox{1.5cm}{$75.62$} &\pbox{1.5cm}{$72.34$} & \pbox{1.5cm}{$74$} & \pbox{1.5cm}{$74$}  \\\hhline{~-----}
      &
      \multicolumn{1}{c|}{DCM \cite{Christop24:online}} & \pbox{1.5cm}{$76.92$} &\pbox{1.5cm}{$74.83$} & \pbox{1.5cm}{$75.85$} & \pbox{1.5cm}{$75$}  \\\hhline{~-----}
      
                                   & \multicolumn{1}{c|}{BCBId} & \pbox{1.5cm}{$71$} & \pbox{1.5cm}{$69.55$} & \pbox{1.5cm}{$70.24$} & \pbox{1.5cm}{$68.56$} \\

                                   \hline
                                   
      \multirow{2}{1.5cm}{NGUYEN et al.\cite{nguyen2017comic}}& \multicolumn{1}{c|}{eBDtheque \cite{guerin2013ebdtheque}} & \pbox{1.5cm}{$79.73$} &\pbox{1.5cm}{$51$} & \pbox{1.5cm}{$62.11$} & \pbox{1.5cm}{$63$}  \\\hhline{~-----} &
      
      \multicolumn{1}{c|}{Manga $109$ \cite{matsui2017sketch}} & \pbox{1.5cm}{$82.78$} &\pbox{1.5cm}{$80.92$} & \pbox{1.5cm}{$81.70$} & \pbox{1.5cm}{$82$}  \\\hhline{~-----}
      
      & 
      
      \multicolumn{1}{c|}{DCM \cite{Christop24:online}} & \pbox{1.5cm}{$80.92$} &\pbox{1.5cm}{$77.85$} & \pbox{1.5cm}{$79.32$} & \pbox{1.5cm}{$77$}  \\\hhline{~-----}
      
                                   & \multicolumn{1}{c|}{BCBId} & \pbox{1.5cm}{$70.11$} & \pbox{1.5cm}{$68.53$} & \pbox{1.5cm}{$69.24$} & \pbox{1.5cm}{$67$} \\

                                   \hline                            
                                   
\multirow{2}{1cm}{Yolo. \cite{redmon2016you}} & \multicolumn{1}{c|}{eBDtheque \cite{guerin2013ebdtheque}} & \pbox{1.5cm}{$60.71$} &\pbox{1.5cm}{$52.76$} & \pbox{1.5cm}{$56.44$} & \pbox{1.5cm}{$58$}  \\\hhline{~-----} &
          \multicolumn{1}{c|}{Manga $109$ \cite{matsui2017sketch}} & \pbox{1.5cm}{$35.38$} &\pbox{1.5cm}{$32.92$} & \pbox{1.5cm}{$34.56$} & \pbox{1.5cm}{$33$}  \\\hhline{~-----} &
          
           \multicolumn{1}{c|}{DCM \cite{Christop24:online}} & \pbox{1.5cm}{$77.23$} &\pbox{1.5cm}{$69.14$} & \pbox{1.5cm}{$73$} & \pbox{1.5cm}{$71$}  \\\hhline{~-----}
                                   & \multicolumn{1}{c|}{{\em BCBId}} & \pbox{1.5cm}{$40.92$} & \pbox{1.5cm}{$35.63$} & \pbox{1.5cm}{$38.15$} & \pbox{1.5cm}{$40$} \\

                                   \hline

\multirow{2}{1cm}{ YOLO9000 \cite{redmon2017yolo9000}} & \multicolumn{1}{c|}{eBDtheque \cite{guerin2013ebdtheque}} & \pbox{1.5cm}{$79.73$} &\pbox{1.5cm}{$55$} & \pbox{1.5cm}{$65.19$} & \pbox{1.5cm}{$67$}  \\\hhline{~-----} &
          \multicolumn{1}{c|}{Manga $109$ \cite{matsui2017sketch}} & \pbox{1.5cm}{$46.94$} &\pbox{1.5cm}{$42.74$} & \pbox{1.5cm}{$44.70$} & \pbox{1.5cm}{$41$}  \\\hhline{~-----} &
          
           \multicolumn{1}{c|}{DCM \cite{Christop24:online}} & \pbox{1.5cm}{$82.3$} &\pbox{1.5cm}{$77.37$} & \pbox{1.5cm}{$79.74$} & \pbox{1.5cm}{$81$}  \\\hhline{~-----}
                                   & \multicolumn{1}{c|}{{\em BCBId}} & \pbox{1.5cm}{$69.26$} & \pbox{1.5cm}{$71.92$} & \pbox{1.5cm}{$70.56$} & \pbox{1.5cm}{$68$} \\

                                   \hline

         \multirow{2}{1.5cm}{Our method \\(Using \\ Softmax)} & \multicolumn{1}{c|}{eBDtheque \cite{guerin2013ebdtheque}} & \pbox{1.5cm}{$91.23$} &\pbox{1.5cm}{$90.56$} & \pbox{1.5cm}{$90.82$} & \pbox{1.5cm}{$90$}  \\\hhline{~-----}
         &
         \multicolumn{1}{c|}{Manga $109$ \cite{matsui2017sketch}} & \pbox{1.5cm}{$92.34$} &\pbox{1.5cm}{$93.76$} & \pbox{1.5cm}{$93.11$} & \pbox{1.5cm}{$92$}  \\\hhline{~-----}
         
         &
         
         \multicolumn{1}{c|}{DCM \cite{Christop24:online}} & \pbox{1.5cm}{$90.55$} &\pbox{1.5cm}{$91.92$} & \pbox{1.5cm}{$91.14$} & \pbox{1.5cm}{$89$}  \\\hhline{~-----}
                                   & \multicolumn{1}{c|}{BCBId} & \pbox{1.5cm}{$93.14$} & \pbox{1.5cm}{$89.91$} & \pbox{1.5cm}{$91.41$} & \pbox{1.5cm}{$91$} \\

                                   \hline

         \multirow{2}{1.5cm}{Our method \\(Using \\ Sigmoid)} & \multicolumn{1}{c|}{eBDtheque \cite{guerin2013ebdtheque}} & \pbox{1.5cm}{$98.52$} &\pbox{1.5cm}{$97$} & \pbox{1.5cm}{$97.74$} & \pbox{1.5cm}{$96.52$}  \\\hhline{~-----}
         &
         \multicolumn{1}{c|}{Manga $109$ \cite{matsui2017sketch}} & \pbox{1.5cm}{$99.14$} &\pbox{1.5cm}{$98.71$} & \pbox{1.5cm}{$98.82$} & \pbox{1.5cm}{$98$}  \\\hhline{~-----}
         
         &
         
         \multicolumn{1}{c|}{DCM \cite{Christop24:online}} & \pbox{1.5cm}{$98.71$} &\pbox{1.5cm}{$98.23$} & \pbox{1.5cm}{$98.41$} & \pbox{1.5cm}{$97$}  \\\hhline{~-----}
                                   & \multicolumn{1}{c|}{BCBId} & \pbox{1.5cm}{$99$} & \pbox{1.5cm}{$98.55$} & \pbox{1.5cm}{$98.74$} & \pbox{1.5cm}{$98$} \\

                                   \hline

         \end{tabular}}{}
\label{tab:xxxcharacter}

\end{table}
\paragraph{Other datasets}
Some publicly available datasets for comic research are eBDtheque \cite{guerin2013ebdtheque}, Manga $109$ \cite{matsui2017sketch}, DCM dataset \cite{Christop24:online} etc. The eBDtheque \cite{guerin2013ebdtheque} dataset contains annotations of $850$ panels and $1550$ comic characters. Whereas Manga $109$ dataset \cite{matsui2017sketch} provides a total of $21,142$ japanese comic book pages. On the other hand, DCM dataset  \cite{Christop24:online} includes a wide variety of comic book pages from Digital Comic Museum \cite{DigitalC66:online} and their annotations.

\paragraph{Evaluation Metrics}
In our method, the final output is represented as bounding boxes across panels and characters. Therefore, we use Intersection Over Union ({\em IoU}) metric to evaluate the matching between the resulted bounding box along with its ground truth. Suppose, $B_{G}$ denotes the ground truth bounding box and $B_{R}$ denotes the resultant bounding box, then The {\em IoU} overlap between them can be defined as follows: $IoU = {\frac{B_{G} \cap  B_{R}}{B_{G} \cup  B_{R} }}$.

Here, the detected boxes, which have more than $80\%$ {\em IoU} overlap with its corresponding ground truth boxes, are taken as true positive $T_{p}$. However, if a resultant box does not match with any ground truth box, then it is regarded as false positive $F_{p}$. Moreover, if there does not exist any match for a ground truth box, then it is considered as false negative $F_{n}$. The value of IoU in Table-\ref{tab:xxxpanel} and Table-\ref{tab:xxxcharacter} is actually the average of IoU Overlap of all finally detected bounding boxes across panels and character respectively.

The method have evaluated using standard metrics like {\em precision}, {\em recall} and {\em F-measure} defined as follows: a) $Precision = {\frac{T_{p}}{T_{p}+ F_{p}}}$; (b) $Recall = {\frac{T_{p}}{T_{p}+ F_{n}}}$; (c) $F-measure = 2 \times {\frac{Precision \times Recall}{Precision + Recall }}$.

\paragraph{Results in detail}
This work can extract both panels/characters from comic books with significantly high accuracy. Figure \ref{resultmanga} and Figure \ref{resultbengali} show the result of the proposed method using Manga $109$ \cite{matsui2017sketch}, {\em eBDtheque} \cite{guerin2013ebdtheque} and Bengali Comic Image dataset respectively. In our method, we have labelled character as person. The performance of panel detection has compared with others using the publicly available eBDtheque \cite{guerin2013ebdtheque}, Manga $109$ \cite{matsui2017sketch}, DCM \cite{Christop24:online} and {\em Bengali} Comic Image Dataset (see Table \ref{tab:xxxpanel}).

To evaluate the performance of comic character detection, we have compared our method with others \cite{rigaud2015knowledge,qin2017faster,nguyen2017comic,sun2013specific} on eBDtheque\cite{guerin2013ebdtheque}, Manga $109$ \cite{matsui2017sketch}, DCM\cite{Christop24:online} and {\em Bengali} Comic Image Dataset (see Table \ref{tab:xxxcharacter}). It is clear from the quantitative analysis of results that the proposed method outperforms in panel/character segmentation from comic book pages.


\begin{figure}[h]
\begin{center}
\begin{tabular}{ll}

a) \fbox{\includegraphics[width=3.2cm,height=4.2cm]{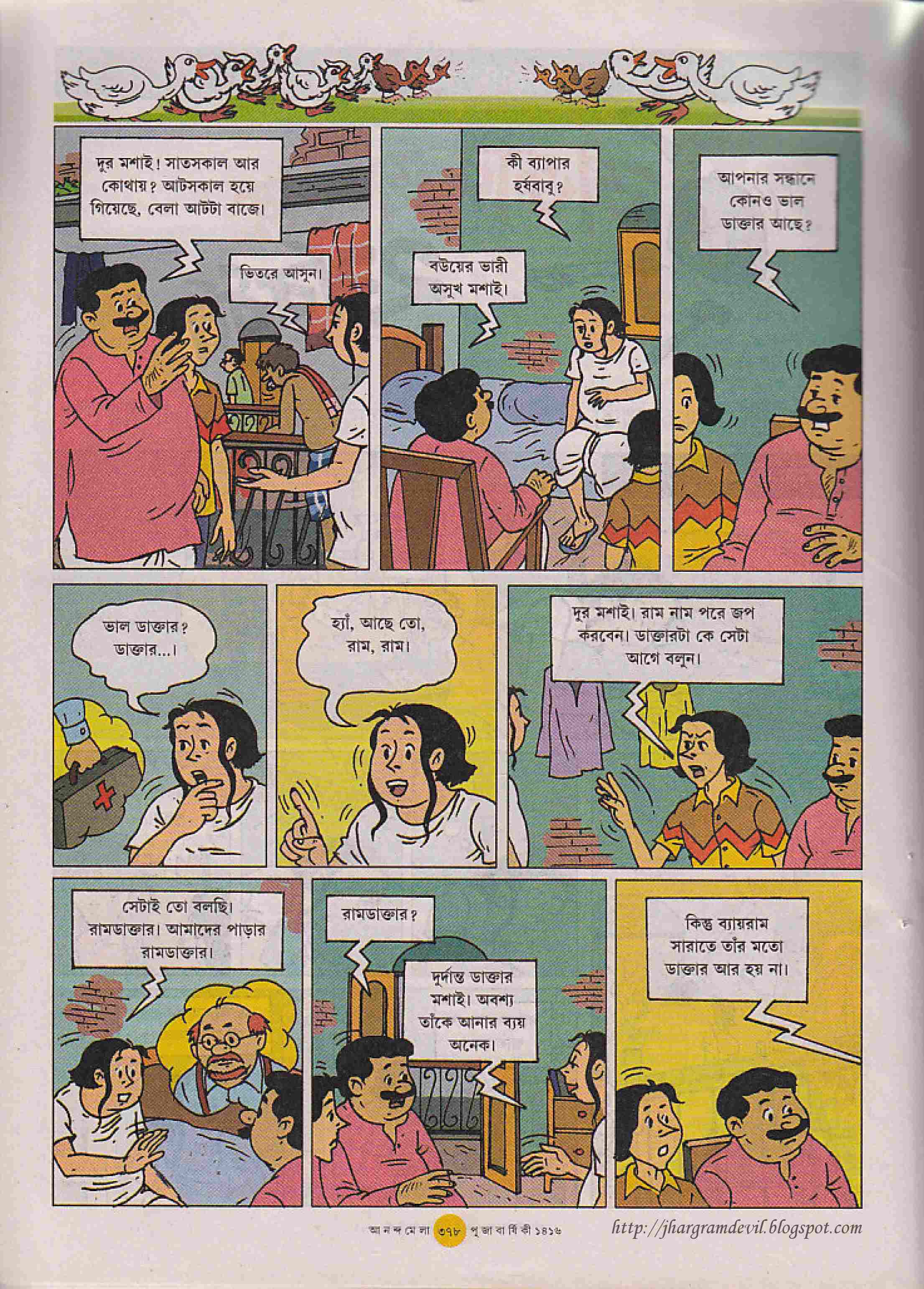}}  & b) \fbox{\includegraphics[width=3.2cm,height=4.2cm]{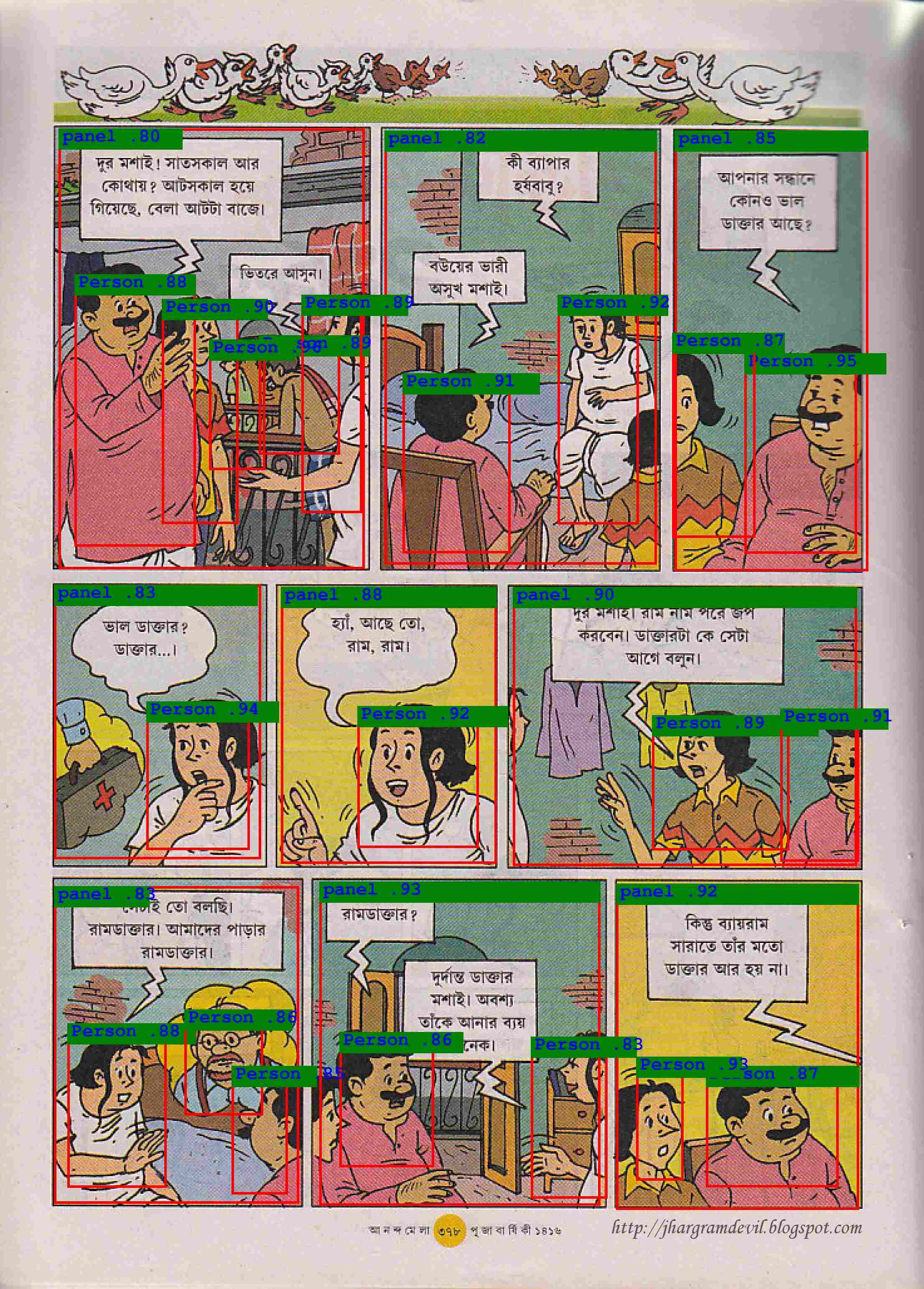}}  \\

\end{tabular}
\end{center}
\caption{Result using proposed method: a) SC1400 dataset; b) Corresponding result.}
	\label{resultbengali}
\end{figure}

\section{Conclusion} \label{ConRemark}

This work successfully extracts and classifies panels/characters from comic book page images. Proper extraction of panels from comic books and then the creation of links among them based on initial order are very necessary to produce digital comics. We are continuously evolving {\em Bengali} comic book page image dataset with a variety of drawing and structural layouts adopted by various {\em Bengali} comic writer and have a plan to make it publicly available.

\begin{spacing}{0.95}
\bibliographystyle{IEEEtran}

\footnotesize{
\bibliography{reference}}
\end{spacing}
%

\end{document}